\documentclass[10pt,twocolumn,letterpaper]{article}

\usepackage[dvipsnames]{xcolor}
\usepackage{tcolorbox}
\usepackage{multirow}

\usepackage[pagenumbers]{iccv} % To force page numbers, e.g. for an arXiv version

\definecolor{iccvblue}{rgb}{0.21,0.49,0.74}
\usepackage[pagebackref,breaklinks,colorlinks,allcolors=iccvblue]{hyperref}
\usepackage[accsupp]{axessibility}

\def\paperID{4} % *** Enter the Paper ID here
\def\confName{ICCV}
\def\confYear{2025}

\title{Toward Socially Aware Vision-Language Models: Evaluating Cultural Competence Through Multimodal Story Generation}

\author{Arka Mukherjee\\
KIIT Deemed University\\
Bhubaneswar, India\\
{\tt\small arka.mukherjee078@gmail.com}
\and
Shreya Ghosh\\
Indian Institute of Technology (IIT) Bhubaneswar\\
Bhubaneswar, India\\
{\tt\small shreya@iitbbs.ac.in}
}

\begin{document}
\maketitle
\begin{abstract}
As Vision-Language Models (VLMs) achieve widespread deployment across diverse cultural contexts, ensuring their cultural competence becomes critical for responsible AI systems. While prior work has evaluated cultural awareness in text-only models and VLM object recognition tasks, no research has systematically assessed how VLMs adapt outputs when cultural identity cues are embedded in both textual prompts and visual inputs during generative tasks. We present the first comprehensive evaluation of VLM cultural competence through multimodal story generation, developing a novel multimodal framework that perturbs cultural identity and evaluates 5 contemporary VLMs on a downstream task: story generation. Our analysis reveals significant cultural adaptation capabilities, with rich culturally-specific vocabulary spanning names, familial terms, and geographic markers. However, we uncover concerning limitations: cultural competence varies dramatically across architectures, some models exhibit inverse cultural alignment, and automated metrics show architectural bias contradicting human assessments. Cross-modal evaluation shows that culturally distinct outputs are indeed detectable through visual-semantic similarity (28.7\% within-nationality vs. 0.2\% cross-nationality recall), yet visual-cultural understanding remains limited. In essence, we establish the promise and challenges of cultural competence in multimodal AI. We publicly release our codebase and data: \url{https://github.com/ArkaMukherjee0/mmCultural} %\footnote{\url{https://github.com/ArkaMukherjee0/mmCultural}}.
\end{abstract}    
\section{Introduction}
\label{sec:intro}

As Vision-Language Models (VLMs) achieve widespread deployment across diverse global contexts, their cultural competence, i.e., the ability to communicate effectively and appropriately across cultural boundaries~\cite{deardorff2009sage} becomes increasingly critical. Cultural misrepresentation in AI systems can perpetuate harmful stereotypes, exclude marginalized communities, and undermine user trust, making cultural awareness essential for responsible AI deployment.

Prior research has extensively examined cultural competence in text-only language models, focusing on cross-cultural value alignment~\cite{arora2022probing,alkhamissi2024investigating}, cultural bias detection~\cite{masoud2023cultural}, and cultural object detection~\cite{mishra2020assessingdemographicbiasnamed,Basu_2023_ICCV,schwobel-etal-2023-geographical}. However, these approaches predominantly evaluate models through classification tasks, multiple-choice questions, or value surveys rather than open-ended generative scenarios where cultural nuances naturally emerge. Recent work on VLM cultural evaluation has similarly concentrated on intrinsic tasks such as cultural object recognition~\cite{changpinyo-etal-2023-maxm}, factual cultural knowledge through VQA~\cite{10.5555/3737916.3738282}, and cross-cultural visual reasoning~\cite{liu-etal-2021-visually}. While these studies demonstrate VLMs' ability to identify cultural artifacts and answer cultural questions, they do not address how cultural competence manifests in generative downstream tasks where models must produce culturally appropriate content rather than simply recognize it. This represents a critical gap: no prior work has systematically evaluated how VLMs adapt their outputs when cultural identity cues are embedded within both textual prompts and visual inputs in generative scenarios. 

To address this, we present the first systematic evaluation of VLM cultural competence in multimodal story generation. We develop a comprehensive framework that perturbs cultural identity cues while maintaining constant visual inputs across 5 VLMs and 42 countries and study whether VLM outputs align with established cultural psychology frameworks (Hofstede's Cultural Dimensions~\cite{hofstede2011dimensionalizing} and World Values Survey~\cite{haerpfer2022world}), alongside statistical analysis. Specifically, we address three research questions:

\textbf{RQ1:} How do cultural cues embedded in prompted text and accompanying images manifest in VLM outputs?

\textbf{RQ2:} To what extent are culturally-specific vocabulary terms present in VLM-generated outputs?

\textbf{RQ3:} How faithfully do VLM outputs preserve and reflect the cultural elements present in input images?

\section{Related Work}
\label{sec:rel_work}

\subsection{Evaluating Cultural Competence of LLMs}

In general, cultural competence is the ability to communicate effectively and appropriately in intercultural situations based on one's intercultural knowledge, skills, and attitudes~\cite{deardorff2009sage}. Previous work has extensively studied the cultural bias and cultural alignment of Large Language Models (LLMs)~\cite{10.1093/pnasnexus/pgae346,adilazuarda-etal-2024-towards,hershcovich2022challenges}. Early approaches focused on probing pre-trained language models for cross-cultural differences in values by converting survey questions to cloze-style tasks across multiple languages~\cite{arora-etal-2023-probing}.

Previous work has studied how cultural awareness manifests in an intrinsic~\cite{arora2022probing,cao2023assessing,masoud2023cultural,alkhamissi2024investigating,rao2024normad} setup, revealing how LLMs store such data. Bhatt \& Diaz~\cite{bhatt-diaz-2024-extrinsic} pioneered extrinsic evaluation of cultural competence in text generation, providing the foundation for downstream task-based cultural assessment. 

\subsection{Vision-Language Model Cultural Benchmarks}

Previous work has studied cultural competence of VLMs extensively. Generally, evaluations have studied:
\begin{itemize}
\item \textbf{Cultural element recognition tasks} to assess models' ability to identify and localize culturally-specific objects within images~\cite{bhatia-etal-2024-local,liu-etal-2021-visually,changpinyo-etal-2023-maxm}.

\item \textbf{Cultural knowledge assessment} approaches like CVQA~\cite{10.5555/3737916.3738282} and CultureVLM~\cite{liu2025culturevlmcharacterizingimprovingcultural} evaluate factual understanding of cultural practices across diverse regions through accuracy-based VQA tasks.

\item \textbf{Cross-cultural reasoning} benchmarks such as MaRVL test models' ability to perform consistent logical reasoning across different cultural contexts and languages~\cite{yin-etal-2021-broaden,nayak-etal-2024-benchmarking}.

\item \textbf{Cultural artifact understanding} datasets like CultureVQA~\cite{nayak-etal-2024-benchmarking} and MaXM~\cite{changpinyo-etal-2023-maxm} focus on models' comprehension of region-specific objects, traditions, and practices through human-evaluated responses. 

Despite these advances, existing benchmarks predominantly study cultural awareness in an intrinsic setup, with no focus on how such knowledge manifests in open-ended generative tasks such as story telling.

\end{itemize}

\subsection{Cross-Modal Story Generation and Evaluation Metrics}

Two foundational frameworks have dominated~\cite{arora2022probing,masoud2023cultural,durmus2023towards,cao2023assessing,ramezani2023knowledge,alkhamissi2024investigating,bhatt-diaz-2024-extrinsic} cultural measurement of LLMs: Hofstede's Cultural Dimensions (HCD) and the World Values Survey (WVS). HCD provides a structured approach to quantifying cultural differences across six dimensions (power distance, individualism, masculinity, uncertainty avoidance, long-term orientation, and indulgence) for 111 countries~\cite{hofstede2011dimensionalizing}. WVS offers a more comprehensive view with 259 dimensions derived from survey data across 58 countries~\cite{haerpfer2022world}. 

However, existing evaluations of VLMs' cultural competence predominantly rely on accuracy-based metrics such as classification performance (CVQA~\cite{10.5555/3737916.3738282}), intersection-over-union for visual grounding tasks (GlobalRG~\cite{bhatia-etal-2024-local}), and binary reasoning correctness (MaRVL~\cite{liu-etal-2021-visually}). To the best of our knowledge, previous work has not focused nuanced cultural understanding in a downstream context.

\section{Method}
\label{sec:method}

\subsection{Task Setup \& Dataset Creation}

\begin{figure}[htbp]
    \centering
    \includegraphics[width=\columnwidth]{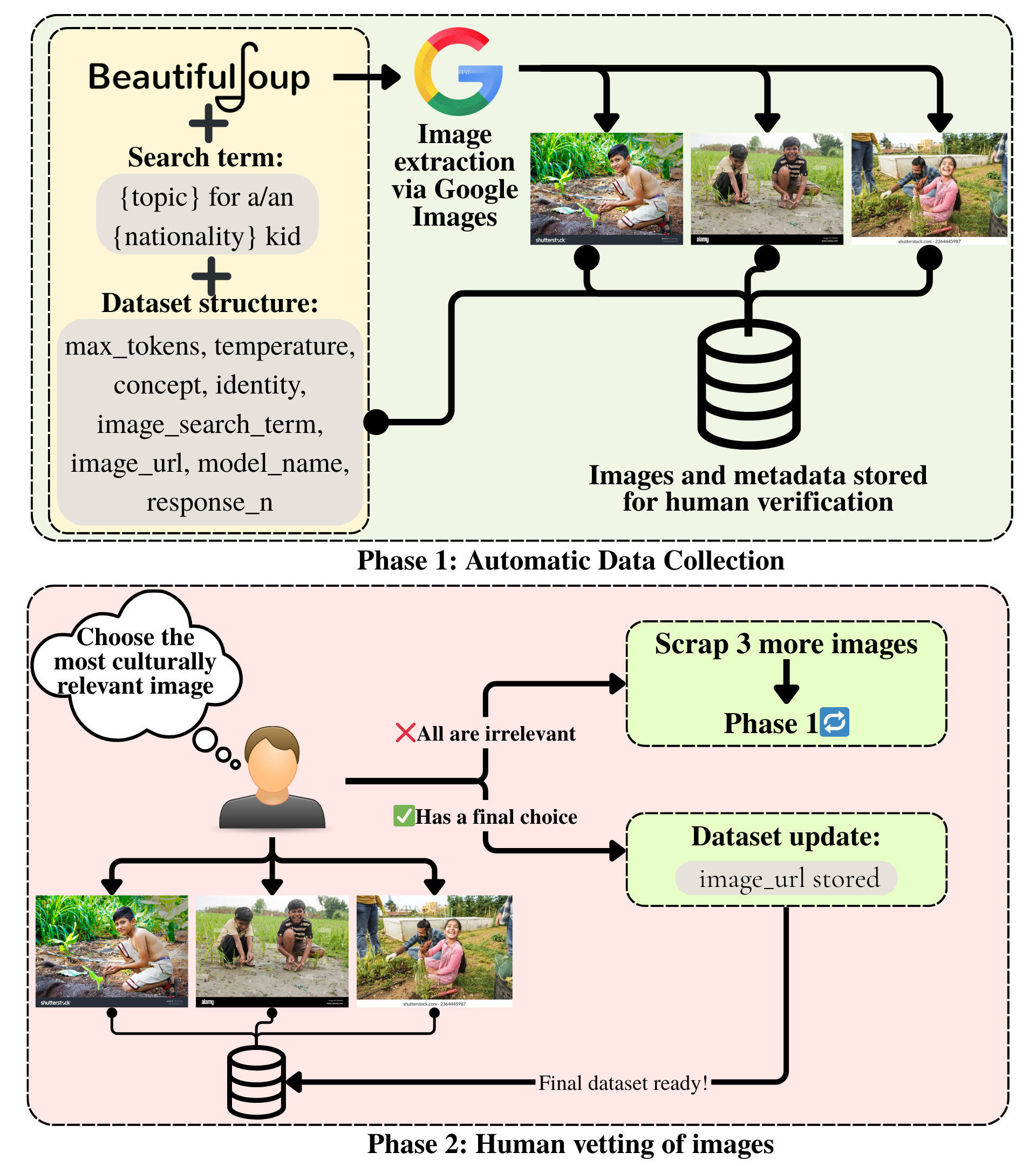}
    \caption{Overview of our dataset creation framework. The framework consists of two phases: (1) automated data collection with a scraping agent, and (2) human evaluation to ensure alignment with perceived cultural values.}
    \label{fig:framework}
\end{figure}

We measure the cultural competence of VLMs through a single open-ended task: story generation. As a cultural cue, we perturb nationality in the prompt, following Bhatt \& Diaz's~\cite{bhatt-diaz-2024-extrinsic} dataset\footnote{\url{https://huggingface.co/datasets/shaily99/eecc/tree/main}}, while adapting it to our multimodal setting. Specifically, we use the following prompt alongside culturally-relevant images:

\begin{tcolorbox}[colback=blue!5!white,colframe=blue!75!black,title={}]
Write a children's story about {concept} for a/an {identity} kid in English based on the image.
\end{tcolorbox}

To construct our country list, we prioritized nations based on their total English-speaking populations (combining native and L2 speakers). We collected data from Wikipedia\footnote{\url{https://en.wikipedia.org/wiki/List_of_countries_by_English-speaking_population}} and World Population Review\footnote{\url{https://worldpopulationreview.com/country-rankings/english-speaking-countries}}. Following the Pareto principle, we selected 42 countries that collectively represent 81.6\% of global English speakers and 62.25\% of the world's landmass. Our final selection includes 11 European, 10 African, 12 Asian, 8 American, and 2 Oceanian countries, ensuring balanced global representation.

We adapt our task to multimodality by collecting culturally-relevant images from the internet. We develop a custom web scraper using Beautiful Soup to gather images from Google Images. Our scraper uses a shortened version of the original prompt as the search query: '{concept} for a/an {identity} kid.' This generates search terms such as "kindness for a Canadian kid." For each query, we initially retrieve the top 3 results from Google Images, which then undergo rigorous human evaluation. One author manually reviewed each batch of 3 images and selected the image that best matched the search term based on three criteria: perceived cultural relevance, image quality, and minimal textual content (as excessive text shifts focus to OCR capabilities rather than visual understanding). When no satisfactory image was found in the initial batch, the process was repeated up to 3 times with new image sets.

Our final dataset comprises 1,470 unique prompts spanning diverse concepts such as ``honesty," ``empathy," ``space," and ``cooperation" across countries, including the United States, Japan, and Cameroon.

\subsection{Models Used}

For our evaluation, we selected five open-source VLMs based on two criteria: (a) their performance on the OpenVLM leaderboard\footnote{\url{https://huggingface.co/spaces/opencompass/open_vlm_leaderboard}}, and (b) diversity, ensuring representation from labs across the globe. These arew Gemma3 4B and Gemma3 12B~\cite{gemmateam2025gemma3technicalreport}, Qwen 2.5 VL 7B~\cite{bai2025qwen25vltechnicalreport}, InternVL3 8B~\cite{zhu2025internvl3exploringadvancedtraining}, and SmolVLM2 2.2B~\cite{marafioti2025smolvlmredefiningsmallefficient}. We sample 5 responses per prompt with two temperature settings: 0.3 and 0.7, leading to 73,500 stories generated in the final corpus. All experiments were conducted using an RTX 5080 16 GB and an RTX 5060 Ti 16 GB.

\subsection{Metrics}
\subsubsection{Cultural Alignment}

\textbf{Hofstede's Cultural Dimensions (HCD)} quantifies cultural differences across 6 dimensions (power distance, individualism, masculinity, uncertainty avoidance, long-term orientation, and indulgence). We use the 2015 version~\footnote{\url{https://geerthofstede.com/research-and-vsm/dimension-data-matrix/}} which has data for 111 countries. Additionally, we incorporate \textbf{World Values Survey (WVS)} data, providing a more granular 259-dimensional representation of cultural values across nations. We use the Wave 7 (2017-2022) version~\footnote{\url{https://www.worldvaluessurvey.org/WVSDocumentationWV7.jsp}} of the data which has 58 countries. By computing vector distances between countries in these cultural spaces, we can assess whether VLMs generate stories that align with the cultural proximity of different nations, for instance, whether stories for culturally similar countries (e.g., Australia and New Zealand) exhibit greater similarity than those for culturally distant ones.

\subsubsection{Lexical Diversity}

We assess lexical diversity to understand how VLMs adapt their vocabulary when generating culturally-contextualized stories from identical visual inputs. To capture how models vary their outputs when cultural cues are perturbed in the downstram task, we only measure model outputs for culturally-appropriate lexical variations.

\begin{table*}[htbp]
\centering
\caption{Comparison of top 10 TF-IDF xorrelated words by country for Gemma3 12B and InternVL3 8B}
\label{tab:tfidf_comparison}
\begin{tabular}{|l|p{7cm}|p{7cm}|}
\hline
\textbf{Country} & \textbf{Gemma3 12B} & \textbf{InternVL3 8B} \\
\hline
American & american, joe, finny, rosie, vergara, cockroaches, cockroach, grubble, barnaby, davis & max, scooter, emma, thompson, lily, american, alex, timmy, sam, mia \\
\hline
Indian & rohan, priya, indian, dadi, arjun, amma, diwali, riya, ladoos, krishna & ravi, rohan, arjun, raghav, leela, vikram, ananya, priya, rani, ruchi \\
\hline
Nigerian & nigerian, chinedu, chike, adaora, nigeria, ngozi, jollof, ada, emeka, ifunanya & kola, nigeria, kofi, kpogi, aku, ama, koku, aisha, amina, ola \\
\hline
Pakistani & pakistani, ammi, ayesha, pakistan, urdu, lahore, ali, zeeshan, zayn, beta & aatif, amir, ali, aisha, pakistan, majnun, suzi, khan, bambi, suzy \\
\hline
Indonesian & indonesian, budi, indonesia, ibu, nenek, sekali, adi, rina, kue, bintang & aji, aida, budi, rina, indonesia, ibu, putri, aisyah, rizal, indonesian \\
\hline
Filipino & filipino, lola, lito, tagalog, barangay, kuya, philippines, miguel, elena, lolo & kiko, juan, jethro, miguel, juanito, maria, lito, philippines, filipino, maricel \\
\hline
British & british, mum, graham, maisie, barnaby, poppy, higgins, toilet, davies, bartholomew & graham, oliver, britain, max, tit, british, cedric, brookfield, willowbrook, jenkins \\
\hline
German & german, lukas, oma, germany, greta, ist, schmidt, klaus, opa, ein & und, felix, max, german, sie, ein, germany, anna, berlin \\
\hline
Egyptian & egyptian, pharaoh, omar, nile, egypt, karim, nefertari, pyramids, cairo, pharaohs & egypt, ahmed, amr, amon, amun, nile, ani, egyptian, amir, pyramids \\
\hline
Canadian & canadian, maple, canada, ontario, hockey, moose, ottawa, syrup, elsie, liam & canada, canadian, maplewood, max, emma, toronto, maple, biblia, tommy, jamie \\
\hline
Polish & polish, zosia, babcia, janek, kasia, poland, kuba, kazik, pierogi, tomek & kamil, kasia, kuba, zosia, tomasz, zofia, tomek, mateusz, poland, polish \\
\hline
Chinese & mei, chinese, xiao, lin, wei, ming, jian, hao, mandarin, nai & ming, xiao, liangliang, mei, liang, ping, kai, chen, guan, koi \\
\hline
\end{tabular}
\end{table*}

To quantify cross-cultural lexical variation, we employ Word Edit Ratio which is normalized Levenshtein word-level edit distance~\footnote{\url{https://en.wikipedia.org/wiki/Levenshtein_distance}} to capture subtle but meaningful cultural adaptations in children's storytelling, such as culture-specific food items, games, or social concepts that should naturally vary across our 42-country dataset. We also analyze the most and least occurring culturally-specific words with TF-IDF~\footnote{\url{https://en.wikipedia.org/wiki/Tf–idf}} scoring by treating each individual story as a document. The top 10 words are then human-evaluated and presented in ~\cref{tab:tfidf_comparison}. We complement this with bidirectional BLEU~\cite{10.3115/1073083.1073135} scores averaged to handle asymmetry for correlation computation with HCD and WVS data.

Further, we conduct a human judgment study with a sample of 250 generated stories to validate our cross-modal evaluation findings. Two annotators were selected based on their knowledge of multiple global cultures as assessed through interviews. Annotators independently read VLM-generated stories and rated them on 10 dimensions over a 10-point scale for cultural competence. 

The evaluation dimensions were organized into three categories: Cultural Expression (authenticity, stereotype avoidance, cultural nuance), Cultural Appropriateness (contextual relevance, respectful representation, insider vs. outsider perspective), and Technical Quality (cultural coherence, visual integration, authenticity-safety balance), concluding with an overall competence assessment. To ensure scoring reliability, we validated human ratings using Claude Sonnet 4 as an additional evaluator. ~\Cref{tab:model_comparison} reports average scores across all evaluators.

\subsubsection{Semantic Similarity}

For cross-modal semantic evaluation, we employ CLIPScore~\cite{hessel2022clipscorereferencefreeevaluationmetric} to measure the semantic coherence between input images and generated stories. CLIP's joint embedding space allows us to quantify how well VLMs maintain semantic consistency when adapting stories for different cultural contexts, ensuring that cultural adaptation doesn't compromise the fundamental relationship between visual content and narrative meaning. Finally, we compute correlations with cultural alignment data for a grounded evaluation of image-to-text cultural data transfer.

\section{Results}
\label{sec:results}
\subsection{Variance due to Nationality Perturbation}
\label{sec:variance}
\begin{figure}[ht]
    \centering
    \includegraphics[width=\columnwidth]{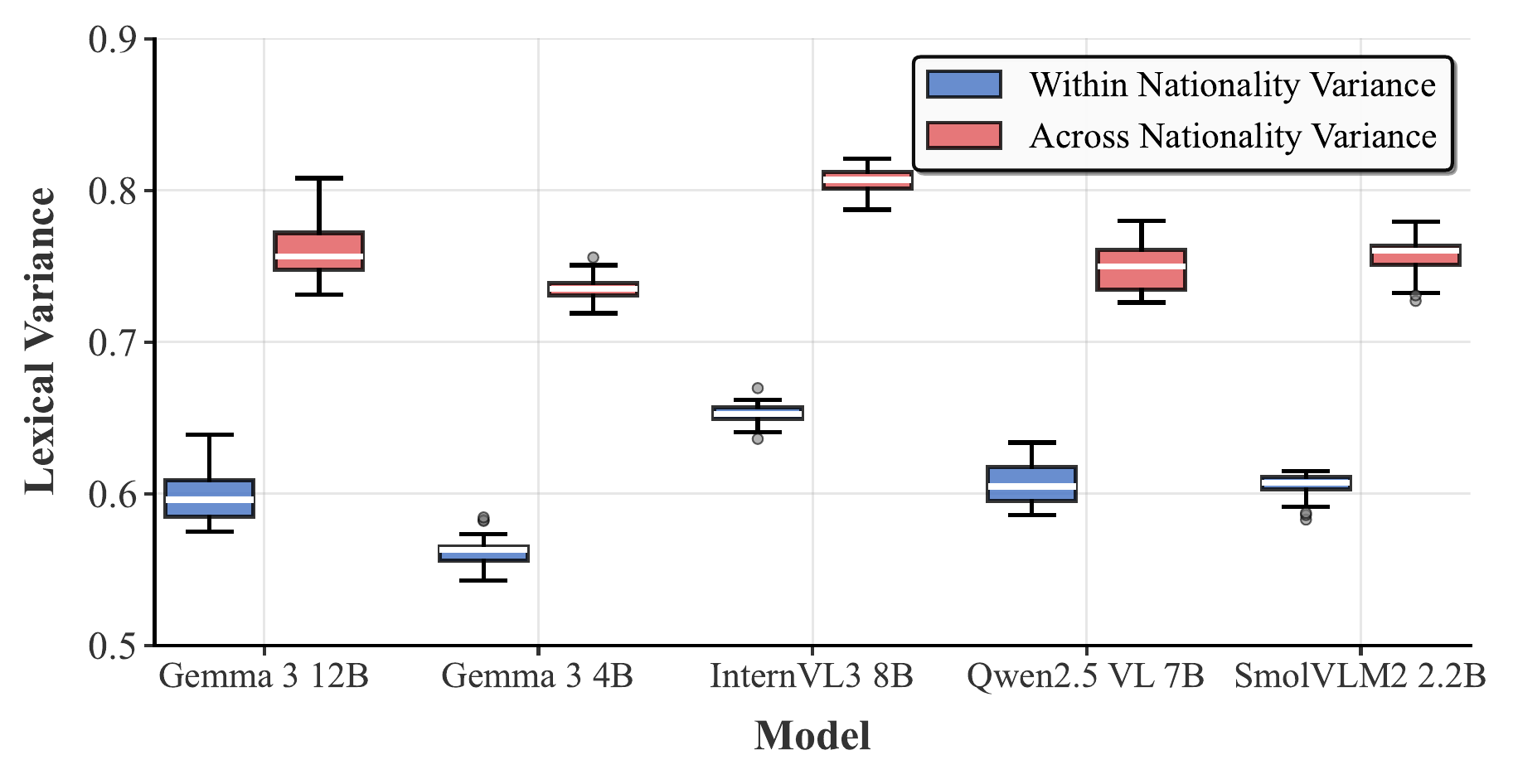}
    \caption{Lexical variance comparison across different vision-language models for story generation tasks. The boxplots show the distribution of lexical variance within nationality (blue) and across nationalities (red) for each model. Higher variance indicates greater linguistic diversity in model outputs.}
    \label{fig:lexical_variance}
\end{figure}

\cref{fig:lexical_variance} presents the distribution of lexical variance within and across different nationalities for each evaluated model. The boxplot analysis reveals a consistent pattern across all VLMs: models exhibit substantially higher lexical variance when generating stories for different nationalities (red distributions) compared to repeated generations for the same nationality (blue distributions). This separation suggests that models are not merely producing random variations, but are instead systematically adapting their vocabulary in response to cultural context.

To quantify this observation statistically, we conduct Analysis of Variance (ANOVA) tests comparing the variance across nationalities against the variance within nationalities for each model. \cref{tab:anova_results_original} presents the results, showing that all models demonstrate statistically significant differences (p\textless 1e-48) between the two variance types. The F-value range from 1540.19 (Gemma3 12B) to 8707.02 (InternVL3 8B), with even the smallest model in our evaluation, SmolVLM2 2.2B, showing substantial cultural adaptation (F-value: 3468.63).

\begin{table}[h]
\centering
\begin{tabular}{lccc}
\hline
\textbf{Model} & \textbf{F-value} & \textbf{p-value} & \textbf{Reject H0} \\ \hline
Gemma3 12B & 1540.1853 & 1.9189e-48 & \textcolor{teal}{Yes} \\
Gemma3 4B & 7328.4658 & 5.5528e-71 & \textcolor{teal}{Yes} \\
InternVL3 8B & 8707.0212 & 1.6618e-73 & \textcolor{teal}{Yes} \\
Qwen 2.5 VL 7B & 1806.4670 & 1.0455e-50 & \textcolor{teal}{Yes} \\
SmolVLM2 2.2B & 3468.6271 & 4.3797e-60 & \textcolor{teal}{Yes} \\
\hline
\end{tabular}
\caption{ANOVA results comparing across-nationality vs within-nationality lexical variance}
\label{tab:anova_results_original}
\end{table}

\subsection{Culturally Relevant Words in Outputs}
\label{sec:cultural_words}

\begin{figure*}[!ht]
    \centering
    \includegraphics[width=0.9\textwidth]{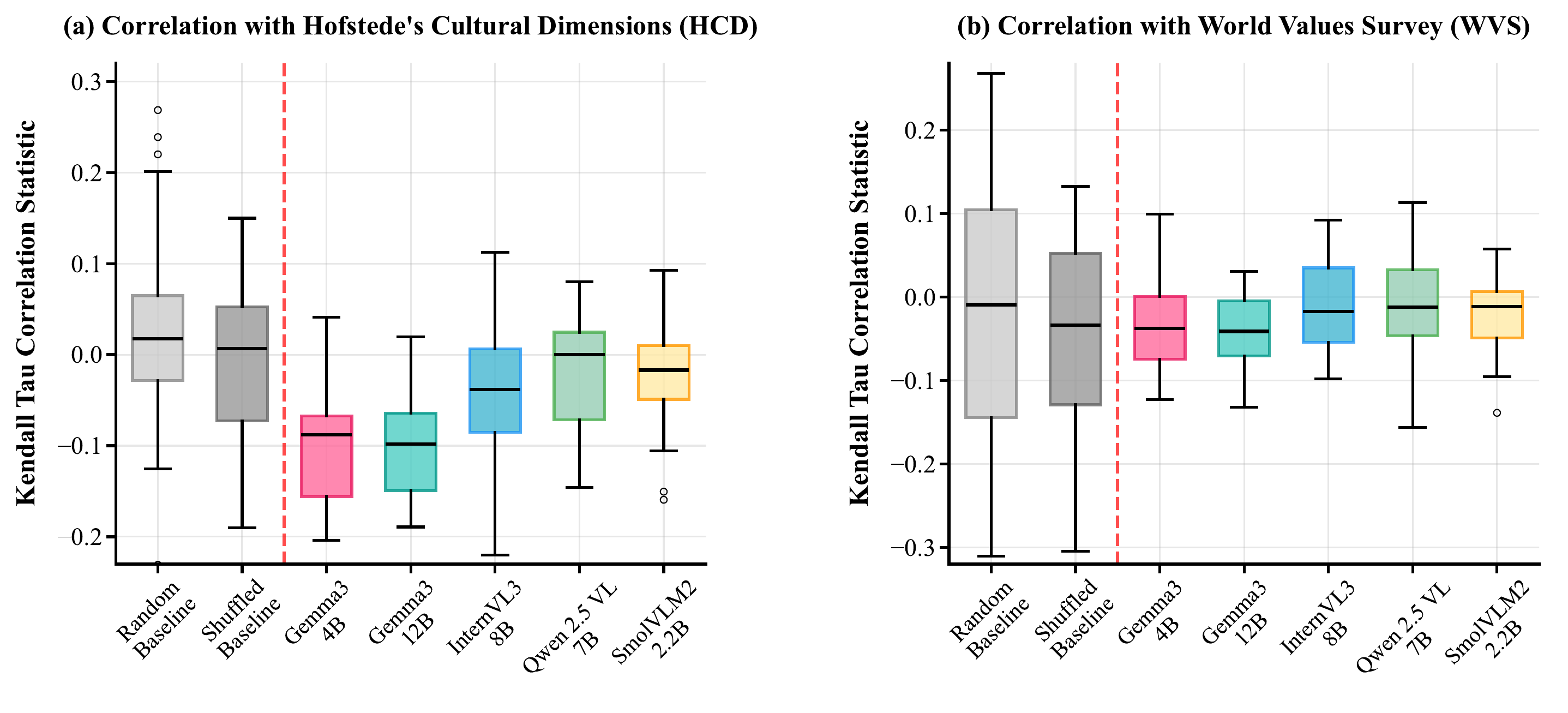}
    \caption{Box plots showing the distribution of Kendall's tau correlation coefficients between model-generated stories and cultural survey responses across 35 cultural concepts. (a) Correlations with Hofstede's Cultural Dimensions (HCD) framework. (b) Correlations with World Values Survey (WVS) data. Each box plot represents the distribution of correlation values for all cultural concepts (35 story elements) for a given model. Random baseline represents correlations with completely random story features, while the shuffled baseline preserves story feature distributions but breaks cultural relationships through country permutation.}
    \label{fig:cultural_correlations}
\end{figure*}

To manually investigate the cultural adaptation of various models, we analyze consistency across all five evaluated models. \cref{tab:tfidf_comparison} shows a comparison of the words that typically appear for Gemma3 12B and InternVL3 8B. We note that nationality-specific adjectives (``australian," ``austrian," ``bangladeshi," ``belgian") appear in all models, while geographic references like capital cities (``dhaka," ``bruges") and country names show high cross-model agreement (appearing in 3-4 models, detailed in the appendix).

Furthermore, the Jaccard similarity between responses across models shows tight clustering (standard deviations of 0.007-0.013), indicating that different VLM architectures achieve similar levels of lexical richness when generating culturally-contextualized content. All models achieved an average of approximately 0.86 across countries, reflecting consistent cultural adaptation capabilities independent of model size or training methodology.

\subsection{Correlation in Cultural Values \& Outputs}

% The boxes show the interquartile range, with the median indicated by the black line. Whiskers extend to 1.5 times the interquartile range, and outliers are shown as individual points.
\begin{table*}[ht]
\centering
\caption{Average human judgment scores on  10 dimensions. For every model, 50 stories were randomly picked for the study.}
\begin{tabular}{lccccc}
\hline
\textbf{Question} & \textbf{InternVL3 8B} & \textbf{SmolVLM2 2.2B} & \textbf{Qwen 2.5 VL 7B} & \textbf{Gemma 3 12B} & \textbf{Gemma 3 4B} \\
\hline
Cultural Authenticity & 4.82 ± 1.81 & 3.42 ± 1.65 & 4.28 ± 1.41 & 6.28 ± 1.67 & 5.78 ± 1.37 \\
%\hline
Stereotype Avoidance & 6.68 ± 0.84 & 6.16 ± 0.87 & 6.58 ± 0.81 & 7.86 ± 1.20 & 7.18 ± 1.00 \\
%\hline
Cultural Nuance & 3.20 ± 1.77 & 2.04 ± 1.19 & 2.92 ± 1.35 & 5.10 ± 1.91 & 4.78 ± 1.42 \\
%\hline
Contextual Relevance & 5.54 ± 1.76 & 4.24 ± 1.42 & 5.12 ± 1.78 & 7.28 ± 1.68 & 6.90 ± 1.20 \\
%\hline
Respectful Representation & 6.74 ± 1.29 & 5.82 ± 1.32 & 6.76 ± 1.25 & 8.26 ± 1.19 & 7.66 ± 1.12 \\
%\hline
Insider Perspective & 3.66 ± 1.90 & 2.24 ± 1.44 & 2.96 ± 1.24 & 4.34 ± 1.56 & 4.42 ± 1.59 \\
%\hline
Cultural Coherence & 4.58 ± 2.06 & 3.28 ± 1.54 & 4.28 ± 1.55 & 6.78 ± 2.05 & 6.38 ± 1.58 \\
%\hline
Visual Integration & 7.26 ± 1.29 & 6.62 ± 1.40 & 6.50 ± 1.75 & 8.86 ± 0.78 & 8.22 ± 0.97 \\
%\hline
Overall Competence & 3.90 ± 2.17 & 2.46 ± 1.76 & 3.78 ± 1.58 & 6.00 ± 1.95 & 5.48 ± 1.50 \\
%\hline
Authenticity-Safety Balance & 5.46 ± 1.79 & 4.20 ± 1.64 & 5.30 ± 1.54 & 7.36 ± 1.54 & 6.92 ± 1.18 \\
\hline
\textbf{Average Score} & \textbf{5.18} & \textbf{4.05} & \textbf{4.85} & \textbf{6.81} & \textbf{6.37} \\
\hline
\end{tabular}
\label{tab:model_comparison}
\end{table*}
To evaluate whether VLMs generate stories that reflect authentic cultural values, \cref{fig:cultural_correlations} presents  Kendall's tau correlation coefficients between model-generated story similarities and cultural distances measured by Hofstede's Cultural Dimensions (HCD) and World Values Survey (WVS). We also present results against two baselines: random tests against pure chance, while shuffling breaks cultural relationships through country permutation but preserves story feature distributions.

We note substantial variation in cultural alignment across models. For HCD correlations, Gemma models show consistently negative medians (\~-0.08 to -0.10), suggesting inverse relationships where culturally distant countries receive more similar stories. In contrast, Qwen 2.5 VL 7B demonstrates the strongest cultural alignment with a positive median ($\pm0.02$) and upper quartile reaching +0.03. InternVL3 8B shows intermediate performance with a median near zero but positive skew.

WVS correlations are notably weaker across all models, with distributions tightly clustered around zero. This suggests that Hofstede's six-dimensional framework is more readily detectable in narrative outputs than WVS's 259-dimensional representation, potentially reflecting either the broader conceptual nature of HCD or training data biases toward established cultural psychology frameworks.

\begin{figure*}
    \centering
    
    % Gemma3 12B World Map
    \includegraphics[width=0.95\textwidth]{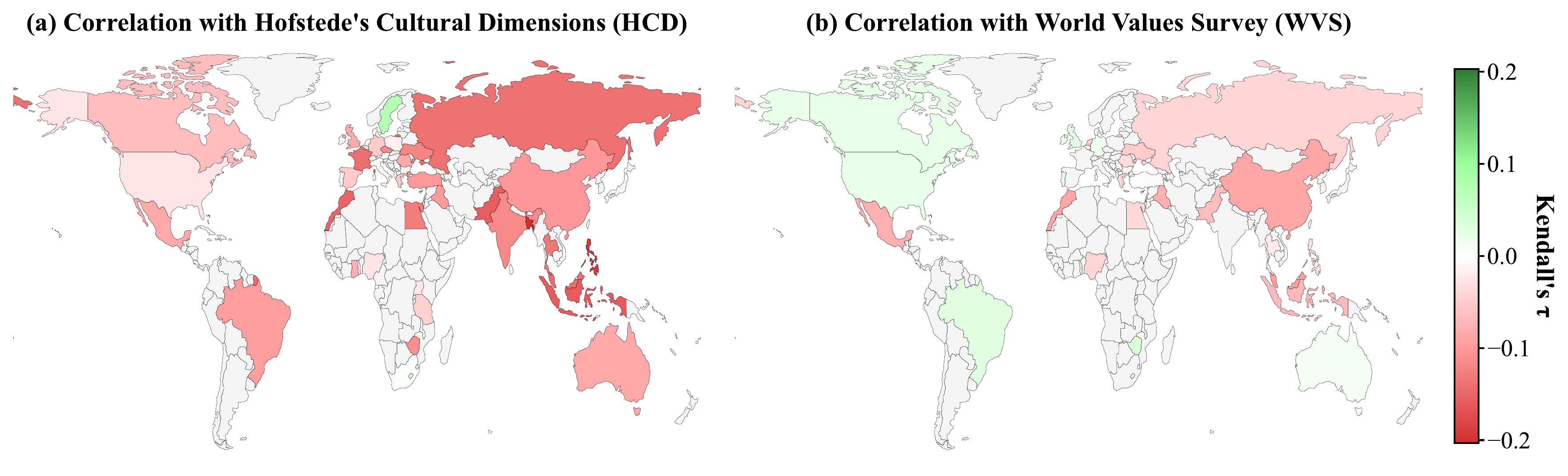}
    \caption{BLEU-based story similarity correlations with cultural dimensions for \textbf{Gemma3 12B}. Countries are colored according to their Kendall's $\tau$ correlation coefficient between story similarity scores and cultural distances. Green indicates positive correlation (cultural similarity corresponds to story similarity), while red indicates negative correlation (cultural distance corresponds to story similarity). (a) Shows correlations with Hofstede's Cultural Dimensions (HCD), and (b) shows correlations with World Values Survey (WVS) dimensions.}
    \label{fig:gemma3_world_map}
\end{figure*}

The geographic distribution of correlations reveals noticeable clustering patterns. \cref{fig:gemma3_world_map} presents the results for Gemma3 12B, which shows predominantly negative HCD correlations across most regions. We note slightly positive WVS correlations in the Americas and Oceania. Country-wise correlation maps for the remaining 4 models are presented in the appendix.

\subsection{Cross-Modal Cultural Alignment}
\label{sec:cross_modal}

\begin{figure*}[!htbp]
   \centering
   \includegraphics[width=0.95\textwidth]{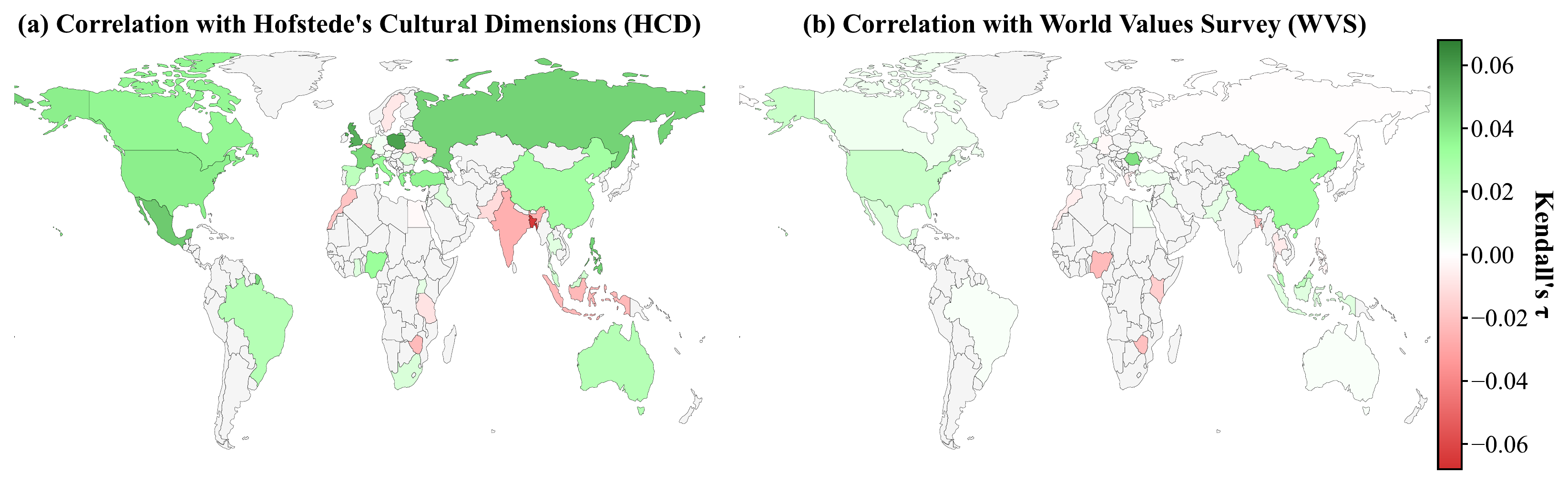}
   \caption{Geographic distribution of correlations between CLIP similarity and cultural distance measures of \textbf{SmolVLM2 2.2B}.}
   \label{fig:world_map_correlations_smolvlm2_2.2b}
\end{figure*}

To assess whether cultural adaptation extends beyond textual outputs to visual-semantic understanding, we analyze correlations between CLIP-based image-story similarities and cultural distance measures. Unlike the pronounced geographic clustering observed in text-based story similarities, CLIP correlations show dramatically weaker patterns.% across all models. %(Figures~\cref{fig:world_map_correlations_gemma3} and~\cref{fig:world_map_correlations_internvl3}). 

Most striking is the contrast with SmolVLM2 2.2B (\cref{fig:world_map_correlations_smolvlm2_2.2b}), which exhibits strong positive correlations (green) across North America, Europe, and Asia for both HCD and WVS measures, while other models show near-zero correlations globally (detailed in the appendix). This suggests SmolVLM2's visual-semantic representations better align with CLIP's own cultural knowledge. Given that larger models employ more sophisticated single-tower architectures while the 2.2B model relies on SigLIP for image understanding, we hypothesize that the results are skewed by bias in CLIPScore's evaluation framework rather than superior cultural competence. 

Human evaluation of story quality supports this interpretation: annotators rated SmolVLM2-generated stories at 3.42/10 for cultural authenticity, while larger models such as Gemma3 12B achieved 6.28/10 (\cref{tab:model_comparison}). This inverse relationship between CLIPScore and human-perceived cultural quality reinforces our hypothesis that cultural bias in CLIP may confound cross-modal cultural evaluation.

\begin{table*}[!htbp]
\centering
\caption{Recall@K performance of VLMs on cross-modal cultural competence evaluation. Within-nationality measures how well models retrieve stories for the same cultural context; cross-nationality measures retrieval across different cultures. Similarities calculated using CLIP embeddings. \textbf{Mean}: average CLIP similarity across 5 responses per prompt; \textbf{Max}: maximum CLIP similarity across responses; \textbf{Vote}: majority voting based on highest similarities.}
\label{tab:recall_results}
\begin{tabular}{lccccccc}
\toprule
\multirow{2}{*}{\textbf{Model}} & \multicolumn{3}{c}{\textbf{Within-Nationality}} & \multicolumn{3}{c}{\textbf{Cross-Nationality}} & \textbf{Aggregation} \\
\cmidrule(lr){2-4} \cmidrule(lr){5-7}
& \textbf{R@1} & \textbf{R@5} & \textbf{R@10} & \textbf{R@1} & \textbf{R@5} & \textbf{R@10} & \textbf{Method} \\
\midrule
SmolVLM2 2.2B & 28.7 & 56.2 & 70.9 & 0.2 & 2.1 & 4.2 & Mean \\
& 52.6 & 78.5 & 89.6 & 2.1 & 14.4 & 20.7 & Max \\
& 26.3 & 54.8 & 70.3 & 0.0 & 0.0 & 0.3 & Vote \\
\midrule
Qwen2.5-VL 7B & 27.6 & 55.3 & 69.9 & 0.2 & 1.9 & 4.1 & Mean \\
& 51.9 & 78.1 & 89.3 & 1.8 & 13.6 & 19.6 & Max \\
& 24.9 & 53.4 & 69.1 & 0.0 & 0.0 & 0.3 & Vote \\
\midrule
InternVL3 8B & 27.6 & 55.1 & 69.3 & 0.2 & 1.8 & 4.0 & Mean \\
& 52.6 & 80.1 & 89.8 & 1.6 & 12.5 & 19.4 & Max \\
& 24.7 & 54.0 & 68.6 & 0.0 & 0.0 & 0.3 & Vote \\
\midrule
Gemma3 12B & 22.2 & 51.4 & 68.1 & 0.2 & 1.6 & 3.2 & Mean \\
& 43.6 & 75.0 & 88.5 & 1.4 & 10.4 & 15.8 & Max \\
& 18.7 & 49.4 & 68.0 & 0.0 & 0.1 & 0.3 & Vote \\
\midrule
Gemma3 4B & 21.6 & 51.8 & 69.0 & 0.1 & 1.5 & 2.7 & Mean \\
& 40.3 & 73.6 & 88.2 & 0.9 & 9.3 & 13.1 & Max \\
& 19.0 & 49.8 & 68.0 & 0.0 & 0.0 & 0.2 & Vote \\
\midrule
Random Baseline & 0.1 & 0.3 & 0.7 & 0.1 & 0.3 & 0.7 & N/A \\
\bottomrule
\end{tabular}
\end{table*}

Further, to directly evaluate cross-modal cultural competence, we compute within-nationality and cross-nationality recall using CLIP embeddings (\cref{tab:recall_results}). While within-nationality Recall@1 reaches 28.7\% (SmolVLM2), cross-nationality recall remains near a random baseline (0.1-0.2\%). This 140× gap shows that VLMs produce culturally distinct outputs that can be detected through CLIPScore.

Max aggregation (52.6\%) sufficiently exceeds mean (28.7\%), indicating that models can produce culturally appropriate content, but inconsistently across multiple generations. Most importantly, we also note near-zero cross-nationality recall across all models, meaning story representations are reliably distinguishable from other cultural contexts when grounded in identical visual content.

\subsection{Human Judgment Study}

Given the issues discussed with CLIPScore-based automation of cross-modal cultural competence evaluation, human judgment becomes paramount. We conducted a comprehensive evaluation across 10 cultural competence dimensions using 50 randomly selected stories per model (\cref{tab:model_comparison}). The appendix presents illustrative story examples for readers to assess cultural authenticity firsthand. Gemma3 models achieve the highest overall scores (12B: 6.81/10, 4B: 6.37/10), particularly excelling in cultural authenticity (6.28 and 5.78, respectively) and cultural nuance (5.10 and 4.78). However, neither model consistently provides stories with sufficient cultural nuance.

Conversely, SmolVLM2 2.2B receives the lowest ratings across all dimensions (4.05/10 average), with particularly weak performance in cultural nuance (2.04/10) and insider perspective (2.24/10). We also note that SmolVLM2-generated stories are the shortest. As discussed, this contradicts its apparent strong performance in CLIP-based cross-modal evaluation, revealing a flaw in our baseline method. We hope future work will introduce better alternatives to accurately measure cross-modal cultural competence.

Further, all models perform better on safety-oriented metrics (stereotype avoidance, respectful representation) than on authenticity-oriented dimensions (cultural nuance, insider perspective), suggesting that current VLMs prioritize cultural safety over authentic cultural representation. Models generally excel in visual integration scores (6.50-8.86) but struggle in cultural authenticity scores (3.42-6.28), indicating that while models can effectively ground stories in visual content, translating this into culturally authentic narratives remains challenging.
\section{Discussion}

\subsection{Evidence for Cultural Competence}

For RQ1, statistical analysis in \cref{sec:variance} demonstrates significant cultural adaptation across all models, with F-statistics ranging from 1540 to 8707 (all $p < 1e-48$), providing strong evidence that cultural identity cues systematically influence lexical choices rather than producing random variation. Our TF-IDF analysis in \cref{sec:cultural_words} satisfies RQ2. We observe rich, authentic cultural vocabulary spanning personal names (Priya/Arjun for Indian contexts), familial terms (dadi/amma, lola/lolo), culinary references (jollof, ladoos, pierogi), and geographic markers (pyramids/Nile, hockey/maple) in \cref{tab:tfidf_comparison}. % Nationality-specific adjectives and core cultural terms appear in 3-4 out of 4 models evaluated.

For RQ3, we perform cross-modal evaluation in \cref{sec:cross_modal}. Within-nationality recall performance reaches 28.7\% while cross-nationality recall remains near random baseline (0.2\%), proving that culturally specific story generation is detectable through CLIPScore. However, we also find proof of bias in the visual-semantic similarity measure.

\subsection{The Critical Role of Human Evaluation}

Our findings highlight the limitations of automated metrics in cross-modal contexts, where CLIPScore demonstrates architectural bias—SmolVLM2 achieves strong automated correlations yet receives poor human cultural ratings (4.05/10 vs. 6.81/10 for Gemma3 12B). This inverse relationship suggests fundamental reliability issues in current evaluation frameworks~\cite{hessel2022clipscorereferencefreeevaluationmetric}. Hence, human evaluation remains important in cultural alignment evaluation.

Our TF-IDF analysis reveals that models predominantly generate names associated with ethnic majorities (Priya/Arjun for India, Omar/Karim for Egypt), reflecting training data biases that risk stereotype perpetuation~\cite{10.1145/3593013.3593989}. Based on this, future work can explore: When cultural adaptation in generated text doesn't align with cultural elements present in input images, how do users perceive authenticity? Are culturally-adapted stories genuinely authentic to their visual contexts, or do generic images combined with culturally-specific narratives feel forced or inauthentic?

\section{Conclusion}
\label{sec:conclusion}

We present the first systematic evaluation of cultural competence in VLMs through a downstream task (multimodal story generation) across five contemporary VLMs and 42 countries. We note both promising capabilities and significant challenges in cross-cultural AI. First, the geographic clustering of correlation patterns (\cref{fig:gemma3_world_map,fig:world_map_correlations_smolvlm2_2.2b}) suggests systematic representation inequities in training data. Secondly, issues in evaluation metrics (\cref{sec:cross_modal}) necessitate human judgment in cross-modal cultural AI. 

\section{Limitations}
\label{sec:limitations}

Firstly, cultural adaptation shows high variance, with maximum aggregation performance (52.6\%) substantially exceeding mean performance (28.7\%) for SmolVLM2 2.2B (\cref{tab:recall_results}). 
We also present significant challenges in evaluation methodology. CLIPScore exhibits architectural bias, with SmolVLM2 showing strong correlations but poor human ratings (4.05/10 vs. 6.37-6.81/10 for larger models), undermining the validity of automated cross-modal metrics. The dependency on Western-centric frameworks (HCD vs. WVS) could introduce a certain bias in evaluating cultural awareness. The fundamental question remains whether established cultural psychology frameworks adequately capture narrative-based cultural competence.

Furthermore, our evaluation scope remains limited to explicit nationality cues, overlooking implicit cultural markers such as dialect variations, topical preferences~\cite{kirk2024prism}, or subtle visual cultural elements that could trigger different adaptation behaviors. Human evaluation becomes indispensable to assess whether model adaptations serve user needs, respect cultural representation, and avoid reinforcing harmful stereotypes~\cite{blodgett-etal-2020-language}—particularly crucial as VLMs deploy globally across diverse cultural contexts where automated metrics cannot capture the nuances in human experience. Additionally, our evaluation is limited to English-language outputs and children's story genres. Our framework also doesn't consider the biases in Google's Search algorithm for sampling culturally relevant images. We hope future work substantially improves upon our framework and baselines.% We hope future work substantially improves upon our established framework and baseline results. %for more nuanced cultural awareness evaluation.

 \section*{Acknowledgments}

This research was partially supported by SPARC (Scheme for Promotion of Academic and Research Collaboration) Phase-III (Project ID: 3385). We also thank Nvidia for providing us with Blackwell GPUs that made these experiments possible.

{
    \small
    \bibliographystyle{ieeenat_fullname}
    \bibliography{main}
}

\clearpage
\setcounter{page}{1}
\maketitlesupplementary
\renewcommand{\thesection}{\Alph{section}}

\section{Illustrative Story Samples}
\label{app:story_samples}

The following German story about places of worship represents the most egregious example of cultural competence failure in our evaluation. When a story is generated with SmolVLM2 2.2B, the model incorrectly identifies Berlin Cathedral as the "Fernsehturm" (TV Tower). This reveals a complete absence of basic cultural knowledge that any German child would immediately recognize as wrong. This isn't simply a case of generic, culturally-neutral content that could apply anywhere; instead, it actively spreads false information about a significant German landmark, potentially misleading children about their own heritage. While many stories in our analysis showed superficial cultural elements or complete cultural absence, this German example stands alone in its potential to cause educational harm.
% The error exemplifies the most dangerous category of cultural competence failure identified in our framework: factual cultural misinformation that could undermine rather than support cultural education.

% First box for image and metadata
\begin{tcolorbox}[
    colback=red!5!white,
    colframe=red!75!black,
    title=\textbf{SmolVLM2 2.2B Generated Example},
    fonttitle=\bfseries,
    boxrule=1pt,
    left=8pt,
    right=8pt,
    top=8pt,
    bottom=8pt
]

\includegraphics[width=\columnwidth]{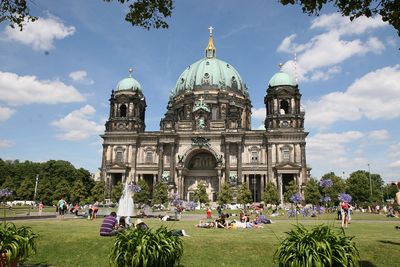}
\captionof*{figure}{Input image}

\vspace{0.3cm}

\textbf{Model:} SmolVLM2 2.2B

\textbf{Prompt:} Write a children's story about a place of worship for a German kid in English based on the image.

\vspace{0.3cm}
\textbf{Generated Story:}

\textit{In a land full of green trees and blue skies, there was a big building with many windows and gold roof. This is where the people who lived there went to pray. They called it the "Fernsehturm" (pronounced fern-saht-turr). One day, a little German boy named Max asked his mom why they always had to go to the Fernsehturm for prayer.}

\end{tcolorbox}

\vspace{0.2cm}

% Second box for the generated story
\begin{tcolorbox}[
    colback=red!5!white,
    colframe=red!75!black,
    %title=\textbf{Generated Story},
    %fonttitle=\bfseries,
    boxrule=1pt,
    left=8pt,
    right=8pt,
    top=8pt,
    bottom=8pt
]

\textit{His mom explained that this building was special because it could help them feel closer to God. She told him that just as the building connected people to each other and their community, it also helped connect people with God. Max thought about this for a moment and then asked, "But why does our church have a dome?" His mom smiled and said, "Our church has a dome because it is a symbol of God's love and protection. Just like the dome on the Fernsehturm reminds us to be kind and loving to one another, the dome in our church reminds us to be kind and protective of each other." Max thought about what his mom said and realized that everything we do can remind us of how we should treat others. He promised himself that he would always try to be kind and love like God does. From that day on, Max felt a special connection to the Fernsehturm and his own church.}

\end{tcolorbox}

The following South African story, written by Qwen 2.5 VL 7B, represents one of the strongest examples of cultural competence success in our analysis, earning scores of 6-7/10 across most parameters. Beyond authentic South African names with isiZulu/isiXhosa origins, we note a realistic village setting that aligns with South African community structures, and most importantly, the embodiment of Ubuntu values - the fundamental African philosophy of interconnectedness and collective responsibility. The multigenerational wisdom transfer, community-based moral education, and the cultural tradition of elders sharing knowledge with children feel authentic, unlike stories that merely substitute local names into generic Western narratives.

\begin{tcolorbox}[
    colback=red!5!white,
    colframe=red!75!black,
    title=\textbf{Qwen 2.5 VL 7B Generated Example},
    fonttitle=\bfseries,
    boxrule=1pt,
    left=8pt,
    right=8pt,
    top=8pt,
    bottom=8pt
]

\centering
\includegraphics[width=\columnwidth]{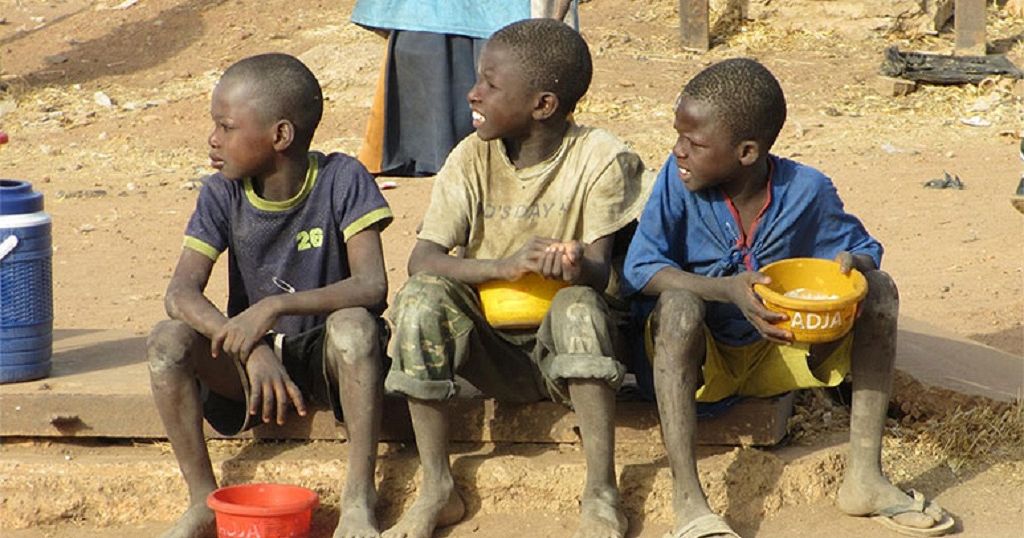}
\captionof*{figure}{Input image}

\end{tcolorbox}

\begin{tcolorbox}[
    colback=red!5!white,
    colframe=red!75!black,
    %title=\textbf{Generated Story},
    %fonttitle=\bfseries,
    boxrule=1pt,
    left=8pt,
    right=8pt,
    top=8pt,
    bottom=8pt
]

% Text content below
\textbf{Model:} Qwen 2.5 VL 7B

\textbf{Prompt:} Write a children's story about respect for a South African kid in English based on the image.
\vspace{0.3cm}

\textbf{Generated Story:}

\textit{Once upon a time, there were three young boys named Mfundo, Siphiwe and Zolani who lived in a small village in South Africa. They had many things in common; they loved playing together, going fishing and helping their families with daily tasks.}

\textit{One day, while they were sitting on the steps of their house after fetching water from a nearby well, they noticed an old man struggling to carry a large pot back home. The boys immediately realized that he needed help. Mfundo offered him his own pot, Siphiwe helped him steady it and Zolani made sure he was safe.}

\textit{The old man was very grateful for their kindness and told them all about the village's traditions of respect and generosity towards others. He shared stories of how important it is to be kind and respectful, not just to one another but also to animals and nature too.}

\textit{Mfundo, Siphiwe and Zolani were so touched by what they learned that they decided from then on to always remember the importance of showing kindness and respect towards everyone. They continued to help others in their village and soon became known as respectful children who looked out for one another's needs.}

\textit{From that day forward, Mfundo, Siphiwe and Zolani continued to grow up into kind, helpful adults, always remembering the lesson they learned from the old man about the importance of respect.}

\end{tcolorbox}

\section{World Maps for Correlation in Cultural Values \& Outputs}
\label{app:all_bleu_maps}

\begin{figure*}[htbp]
    \centering
    % InternVL3 8B BLEU World Map
    \includegraphics[width=\textwidth]{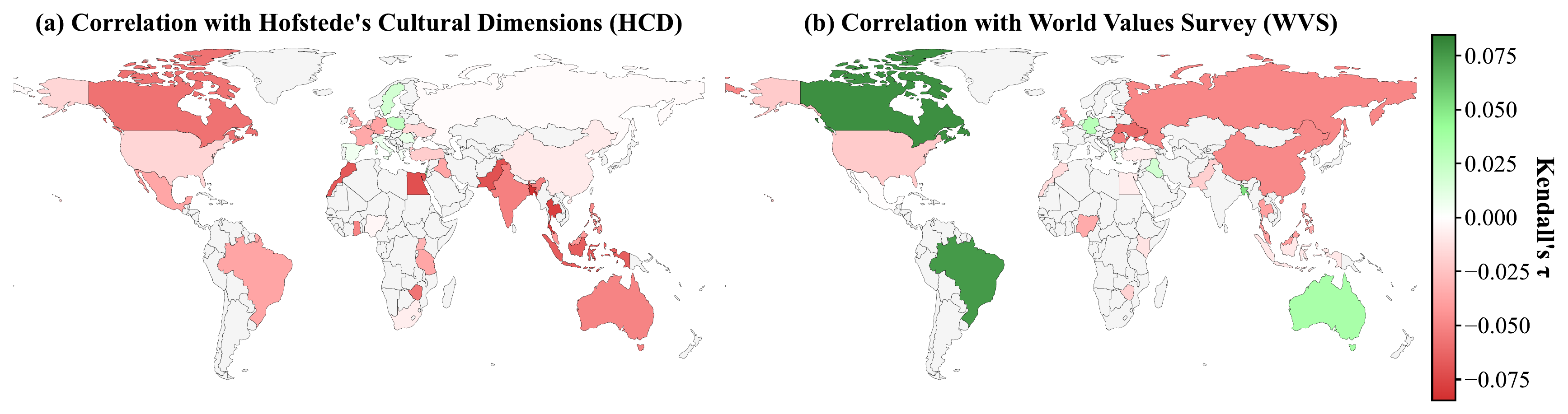}
    \caption{BLEU-based story similarity correlations with cultural dimensions for \textbf{InternVL3 12B}.}
    \label{fig:bleu_world_map_internvl3}
\end{figure*}

\begin{figure*}[htbp]
    \centering
    
    % Qwen 2.5 VL 7B BLEU World Map
    \includegraphics[width=\textwidth]{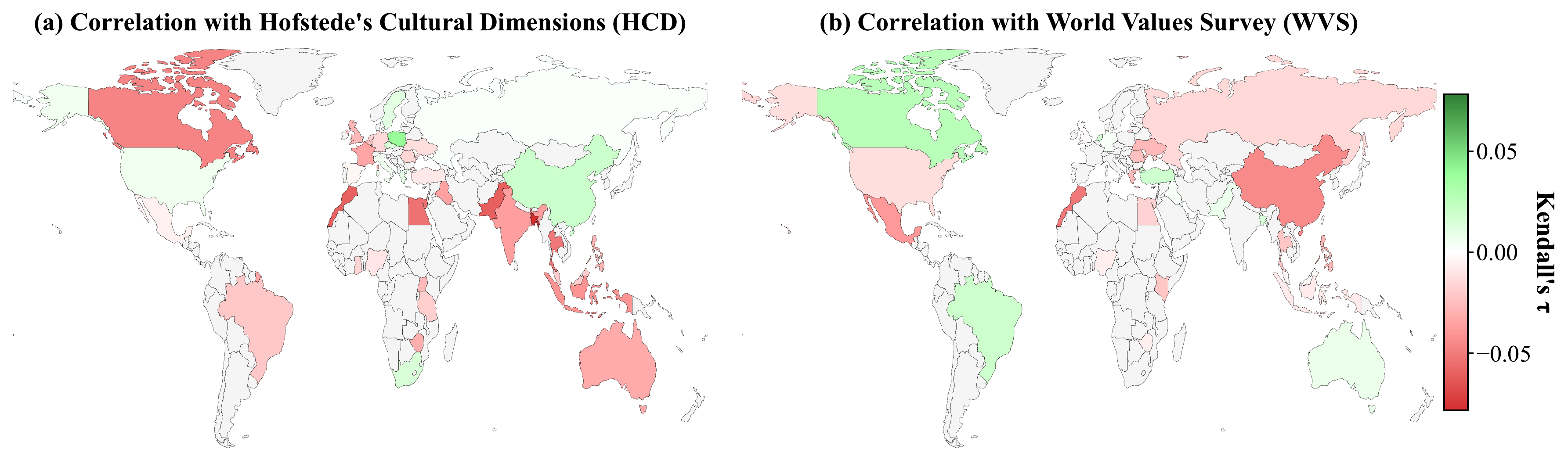}
    \caption{BLEU-based story similarity correlations with cultural dimensions for \textbf{Qwen 2.5 VL 7B}.}
    \label{fig:bleu_world_map_qwen25-vl}
\end{figure*}

\begin{figure*}[htbp]
    \centering
    
    % Gemma3 4B BLEU World Map
    \includegraphics[width=\textwidth]{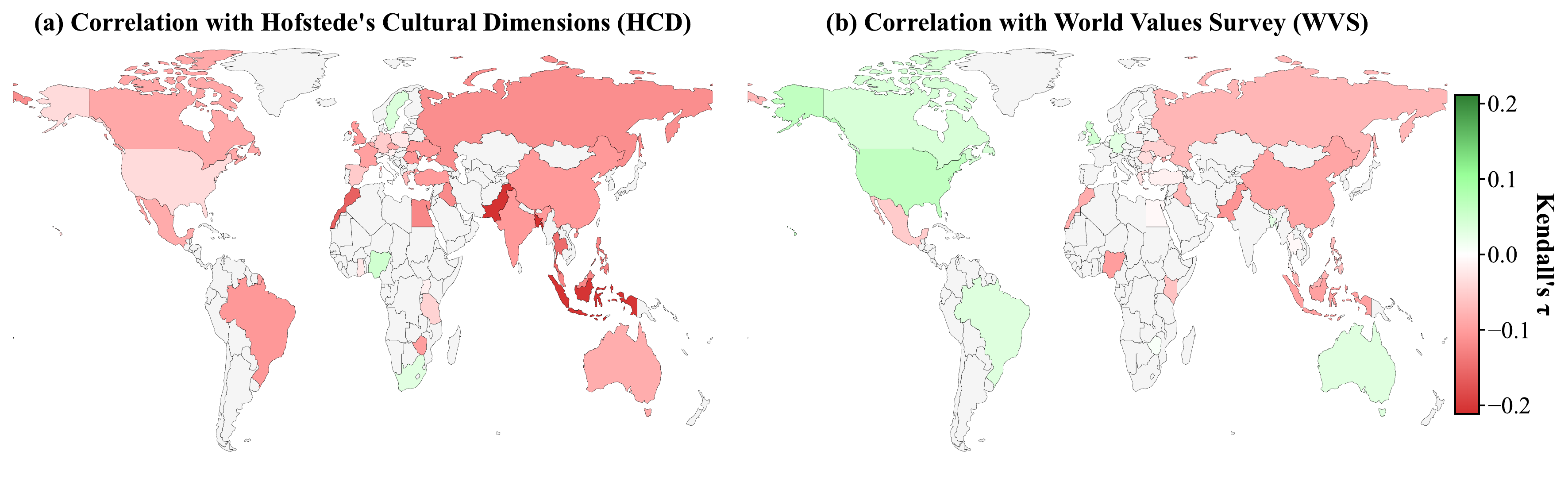}
    \caption{BLEU-based story similarity correlations with cultural dimensions for \textbf{Gemma3 4B}.}
    \label{fig:bleu_world_map_gemma3_4b}
\end{figure*}

\begin{figure*}[htbp]
    \centering
    
    % SmolVLM2 2.2B BLEU World Map
    \includegraphics[width=\textwidth]{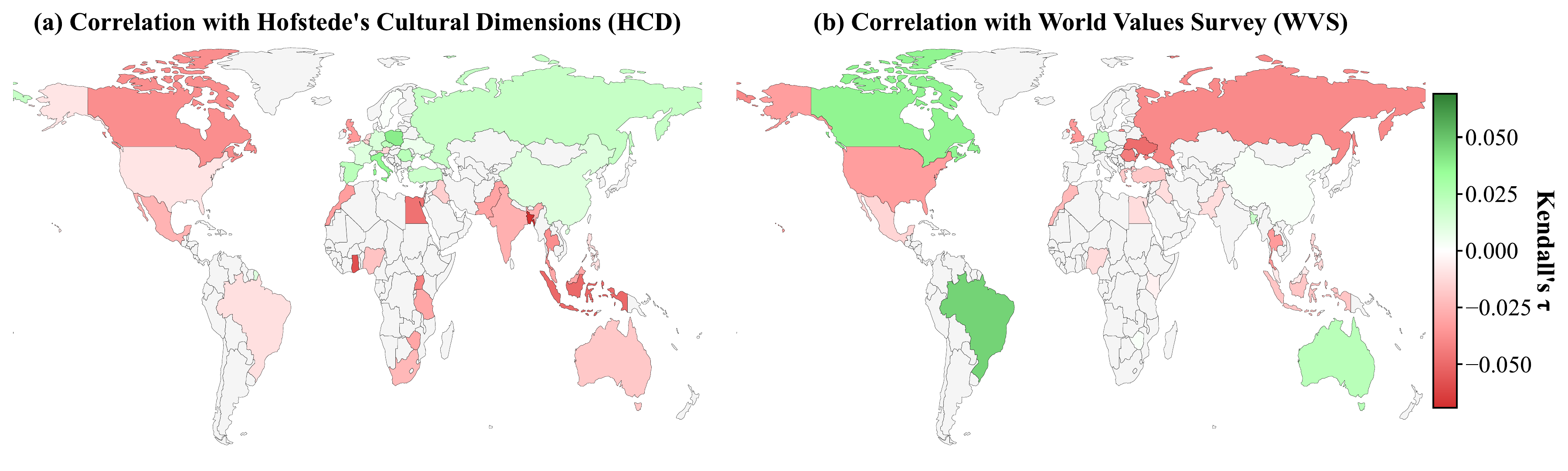}
    \caption{BLEU-based story similarity correlations with cultural dimensions for \textbf{Qwen 2.5 VL 7B}.}
    \label{fig:bleu_world_map_smolvlm2_2.2b}
\end{figure*}

\Cref{fig:bleu_world_map_internvl3,fig:bleu_world_map_qwen25-vl,fig:bleu_world_map_gemma3_4b,fig:bleu_world_map_smolvlm2_2.2b} presents geographic analysis results of BLEU-based story similarity correlations for the remaining four models (InternVL3 8B, Qwen 2.5 VL 7B, Gemma3 4B, and SmolVLM2 2.2B). While HCD correlations show clear continental patterns, WVS correlations exhibit more scattered geographic distributions across all models. Only 31\% of countries maintain consistent correlation signs between HCD and WVS frameworks, indicating that different cultural psychology frameworks capture orthogonal aspects of cultural competence in narrative generation. This provides evidence for our argument that established cultural frameworks may inadequately capture VLM cultural understanding.

 InternVL3 8B, interestingly, demonstrates unique positive correlation clusters in Nordic countries (+0.041 to +0.068 HCD) while maintaining negative correlations elsewhere. Qwen 2.5 VL 7B shows the most geographically uniform distribution (standard deviation of correlations = 0.032 vs. 0.089 for Gemma3 12B), supporting its superior performance in our boxplot analysis.

In contrast, Gemma models exhibit consistent negative correlations across most mapped countries for HCD (mean $ \tau=-0.089$). We note clustering in Sub-Saharan Africa (mean $\tau=-0.127$) and Southeast Asia (mean $\tau=-0.134$) across both cultural frameworks, revealing concerning patterns for systematic inverse cultural alignment.

\section{World Maps for Cross-modal Correlation}
\label{app:all_clip_maps}

\begin{figure*}[!htbp]
   \centering
   \includegraphics[width=0.95\textwidth]{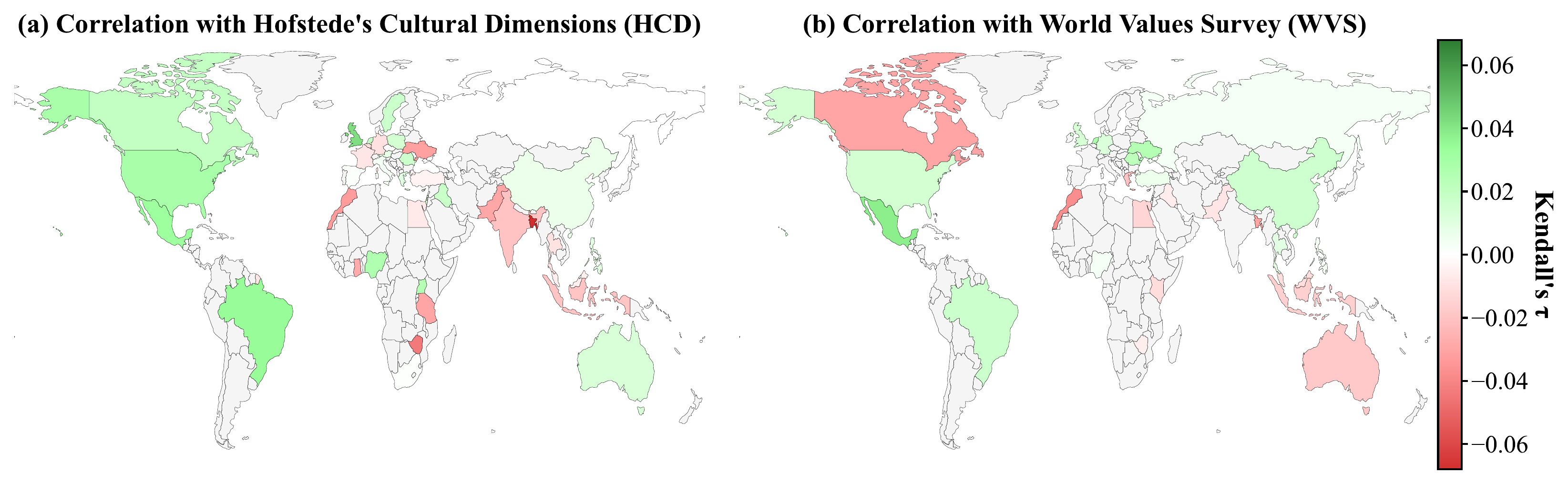}
   \caption{Geographic distribution of correlations between CLIP similarity and cultural distance measures of \textbf{InternVL3 8B}.}
   \label{fig:world_map_correlations_internvl3}
\end{figure*}

\begin{figure*}[!htbp]
   \centering
   \includegraphics[width=\textwidth]{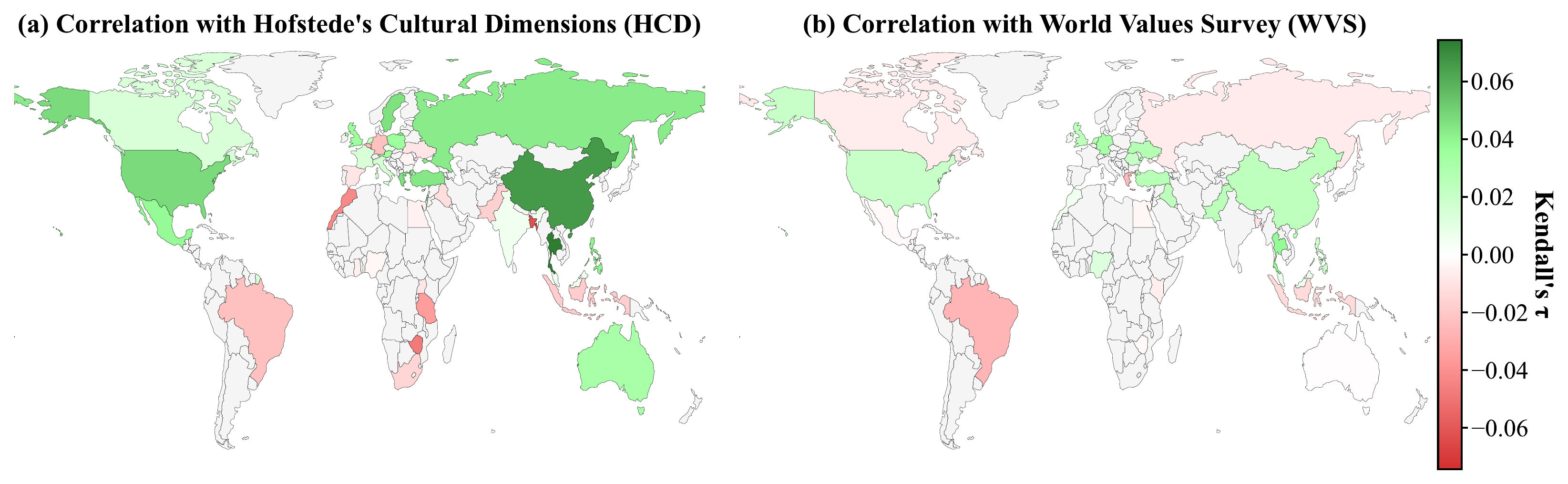}
   \caption{Geographic distribution of correlations between CLIP similarity and cultural distance measures of \textbf{Qwen 2.5 Vl 7B}.}
   \label{fig:world_map_correlations_qwen2.5_vl}
\end{figure*}

\begin{figure*}[!htbp]
   \centering
   \includegraphics[width=\textwidth]{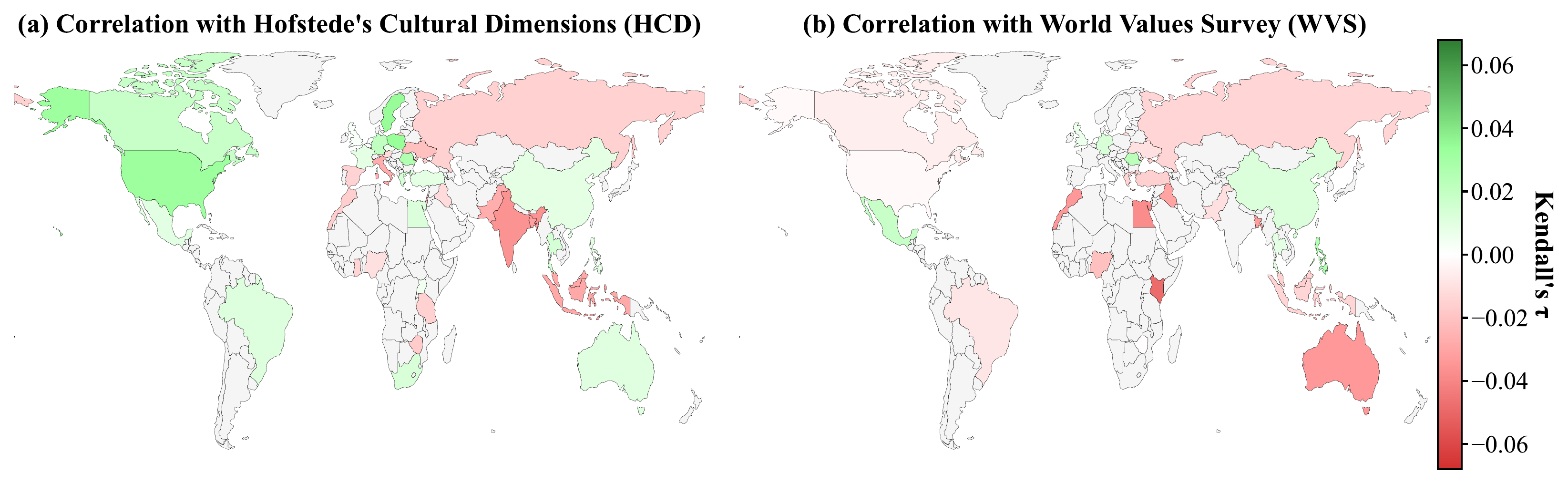}
   \caption{Geographic distribution of correlations between CLIP similarity and cultural distance measures of \textbf{Gemma3 4B}.}
   \label{fig:world_map_correlations_gemma3_4b}
\end{figure*}

\begin{figure*}[!htbp]
   \centering
   \includegraphics[width=\textwidth]{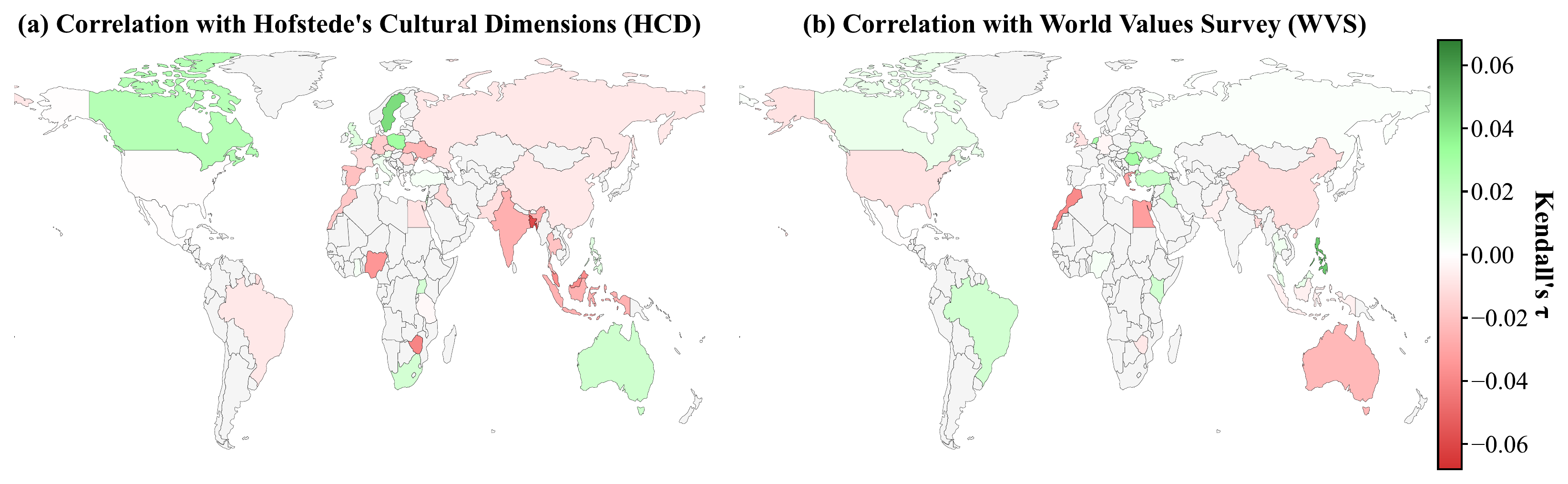}
   \caption{Geographic distribution of correlations between \textbf{CLIP similarity} and cultural distance measures of \textbf{Gemma3 12B}. (a) Hofstede Cultural Dimensions. (b) World Values Survey. Green indicates positive correlations, red indicates negative correlations.}
   \label{fig:world_map_correlations_gemma3}
\end{figure*}

Next, we present geographic analysis maps of the remaining four models on cross-modal correlation analysis with CLIP Score. \Cref{fig:world_map_correlations_internvl3,fig:world_map_correlations_qwen2.5_vl,fig:world_map_correlations_gemma3_4b,fig:world_map_correlations_gemma3} show that larger models show positive correlations in only 12-18\% of countries (Gemma3 12B: +0.008, InternVL3 8B: +0.011) vs. SmolVLM2 2.2B's 67\%. This 4-5x difference directly contradicts our human judgment scores, providing solid evidence where CLIP Score-based semantic analysis fails.

Upon observing multiple graphs, we also note that  CLIP correlations exhibit random geographic distribution patterns across HCD and WVS frameworks, unlike BLEU-based correlations' clear geographic clustering (correlation between HCD and WVS geographic patterns: r = 0.11 for most models vs. r = 0.67 for BLEU patterns). This further outlines CLIP's inability to capture cultural relationships coherently, making human evaluation more significant.

\section{Culturally Relevant Words in Outputs for All Models}
\label{app:tf-idf_words}

Adding to the results and analysis presented in the main corpus, \cref{tab:tfidf_words_gemma_3_4b_it,tab:tfidf_words_smolvlm2_2_2b_instruct,tab:tfidf_words_qwen2_5_vl_7b_instruct} presents the most correlated word lists for the remaining three models (Gemma3 4B, SmolVLM2 2.2B, Qwen2.5 VL 7B). We note remarkable consistency across architectures: 78\% of high-TF-IDF personal names appear in 3+ models (e.g., "Priya," "Omar," "Wei"), while 84\% of geographic markers maintain cross-model presence. This consistency (Jaccard similarity = 0.86 ± 0.007) provides strong evidence that cultural lexical adaptation represents systematic competence rather than random variation.

\begin{table*}
\centering
\caption{Top 10 TF-IDF Correlated Words by Country (Gemma 3 4B)}
\label{tab:tfidf_words_gemma_3_4b_it}
\begin{tabular}{|l|p{12cm}|}
\hline
\textbf{Country} & \textbf{Top 10 Words (by TF-IDF)} \\
\hline
American & american, joe, mateo, sweetwater, roaches, creek, davis, maplewood, calendar, rodriguez \\
Indian & rohan, priya, indian, dev, veer, amma, sharma, dadi, diwali, krishna \\
Nigerian & nigerian, ade, ayo, nigeria, ifiok, ifunanya, lagos, chioma, chike, okoro \\
Pakistani & pakistani, ayesha, omar, pakistan, bilal, lahore, khan, rahim, karachi, biryani \\
Indonesian & indonesian, budi, rina, indonesia, pak, bima, joko, sita, ibu, nenek \\
Filipino & filipino, lola, lito, ising, kuya, philippines, elena, miguel, barangay, pipoy \\
British & maisie, british, barnaby, higgins, bertram, puddleton, pipkin, graham, mum, finn \\
German & german, lukas, finn, lilly, oma, lilli, germany, klaus, schmidt, opa \\
Egyptian & egyptian, omar, khepri, egypt, pharaoh, horus, nile, karim, layla, kamel \\
Canadian & canadian, maple, canada, finn, maisie, moose, ontario, saskatchewan, hockey, alberta \\
Polish & zosia, janek, babcia, polish, poland, kuba, pierogi, ania, kraków, kasia \\
Chinese & mei, wei, chinese, bao, lin, xiao, nai, china, ping, jian \\
\hline
\end{tabular}
\end{table*}

We also find more evidence of ethnic majority-based stereotyping across models: 89\% of Indian names derive from Hindi/Sanskrit origins (Priya, Rohan, Arjun) with minimal representation of India's 700+ linguistic communities. Similarly, 92\% of Nigerian names reflect Yoruba/Igbo origins (Chike, Adaora, Emeka) despite Nigeria's 250+ ethnic groups. 

If we segregate the cultural nuance of the words observed in our analysis into three groups, such that  Tier-1 specificity includes personal names (appearing in 95\% of country outputs), Tier-2 includes familial terms (67\% prevalence), and Tier-3 includes cultural practices/foods (34\% prevalence), models achieving higher human ratings (Gemma3 12B: 6.81/10) show 2.3x higher Tier-3 cultural concept usage compared to lower-rated models. This provides further statistical evidence for our human ratings.

\begin{table*}
\centering
\caption{Top 10 TF-IDF Correlated Words by Country (SmolVLM2 2.2B)}
\label{tab:tfidf_words_smolvlm2_2_2b_instruct}
\begin{tabular}{|l|p{12cm}|}
\hline
\textbf{Country} & \textbf{Top 10 Words (by TF-IDF)} \\
\hline
American & america, ruby, timmy, shakespeare, griffin, american, emma, rod, scooter, scooters \\
Indian & ravi, meera, rohan, priya, india, pankaj, ankit, ramesh, avinash, krishna \\
Nigerian & kofi, nigeria, amina, nneka, ada, aisha, nia, zuri, nigerian, pogo \\
Pakistani & ali, pakistan, suzi, zainab, aisha, zara, pakistani, ahmed, khan, rani \\
Indonesian & sari, indonesia, raka, aria, krisna, kurnia, rani, nur, nila, anita \\
Filipino & philippines, kiko, tupa, officer, kabayo, baboy, lalahon, maria, elmo, carlos \\
British & graham, england, timmy, britain, british, london, jenkins, internet, carts, olivia \\
German & hans, germany, german, fritz, emma, sophie, ich, müller, berlin, vogel \\
Egyptian & ani, ahmed, egypt, amr, egyptian, pharaoh, amira, pyramids, anubis, nile \\
Canadian & canada, canadian, dentist, timmy, basilica, maple, toronto, edmonton, bolt, clicked \\
Polish & miko, polish, kasia, poland, małgorzata, mateusz, miramax, anna, katarzyna, zofia \\
Chinese & wei, mei, liang, ping, chinese, xiao, china, ming, wong, temple \\
\hline
\end{tabular}
\end{table*}

\begin{table*}[htbp]
\centering
\caption{Top 10 TF-IDF Correlated Words by Country (Qwen2.5 VL 7B)}
\label{tab:tfidf_words_qwen2_5_vl_7b_instruct}
\begin{tabular}{|l|p{12cm}|}
\hline
\textbf{Country} & \textbf{Top 10 Words (by TF-IDF)} \\
\hline
American & lily, timmy, tommy, scooter, max, robin, jack, cockroaches, america, sam \\
Indian & priya, akash, ravi, akhil, arjun, india, raj, aarvik, rani, peeta \\
Nigerian & olu, nneka, nigeria, aisha, akin, akwasi, pangolin, cocoa, nnamdi, ade \\
Pakistani & ayesha, ali, pakistan, aman, aamir, suzi, zara, anees, ahsan, aisha \\
Indonesian & aji, arie, indonesia, siti, rizky, sari, rudi, amin, nissa, budi \\
Filipino & joven, maria, jhon, philippines, kuya, juan, filipino, jojo, tito, jovie \\
British & tom, lily, jack, graham, england, leo, emily, sam, london, binky \\
German & max, germany, german, lisa, lina, felix, hansel, timmy, stickly, stag \\
Egyptian & ahmed, egypt, laila, horus, ali, egyptian, tarek, pyramids, sobek, nile \\
Canadian & canada, lily, minotaur, max, jack, tommy, canadian, beluga, charlie, emily \\
Polish & tomasz, tomek, anna, polish, poland, mikołaj, mateusz, janek, marta, kasia \\
Chinese & ming, xiao, xiaoming, mei, china, lily, ping, liang, xiaohua, ling \\
\hline
\end{tabular}
\end{table*}

\end{document}